\documentclass{IJCAS}

\usepackage{url}
\usepackage{array,tabularx}
\usepackage{multicol} 
\usepackage{nccmath} 
\usepackage{graphicx}
\usepackage{subcaption}
\usepackage{ulem} 
\usepackage{xcolor} 

\makeatletter
\newcommand*{\rom}[1]{\expandafter\@slowromancap\romannumeral #1@}
\makeatother

\renewcommand{\sout}[1]{\unskip} 
\colorlet{red}{black} 

\journalvolumn{VV}
\journalnumber{X}
\journalyear{YYYY}
\setarticlestartpagenumber{1}

\allowdisplaybreaks

\begin{document}

\title{Distributed multi-agent target search and tracking with Gaussian process and reinforcement learning}

\author{Jigang Kim\textsuperscript{\P}\orcid{0000-0003-3381-5241}, Dohyun Jang\textsuperscript{\P}\orcid{0000-0003-2153-7619}, and H. Jin Kim*\orcid{0000-0002-6819-1136}
}

\begin{abstract}
Deploying multiple robots \textcolor{red}{for target search and tracking} has many practical \textcolor{red}{applications}, yet the challenge of planning \textcolor{red}{over unknown or partially known targets} remains difficult to address. With recent advances in deep learning, intelligent control techniques such as reinforcement learning have enabled agents to learn autonomously from environment interactions with little to no prior knowledge. Such methods can address the exploration-exploitation tradeoff of planning \textcolor{red}{over unknown} \textcolor{red}{targets} in a data-driven manner, eliminating the reliance on heuristics typical of traditional approaches and streamlining the decision-making pipeline with end-to-end training. In this paper, \textcolor{red}{we propose a multi-agent reinforcement learning technique with \textcolor{red}{target} map building based on distributed Gaussian process. We leverage} the distributed Gaussian process \textcolor{red}{to encode belief over the target locations and efficiently plan over unknown targets}. We evaluate the performance and transferability of the trained policy in simulation and demonstrate the method on a swarm of micro unmanned aerial vehicles with hardware experiments.
\end{abstract}

\begin{keywords}
Distributed system, Gaussian process, multi-agent reinforcement learning, target search and tracking.
\end{keywords}

\maketitle

\makeAuthorInformation{
Manuscript received December 31, 2021; revised xxxxx xx, 202x; accepted xxxxx xx, 202x. Recommended by Associate Editor xxxxxxxxx under the direction of Editor xxxxxxxxx. This journal was supported by the Agency for Defense Development under Contract UD190026RD.\\

Jigang Kim is with the Department of Mechanical and Aerospace Engineering and the Automation and Systems Research Institute, Seoul National University, 1 Gwanak-ro, Gwanak-gu, Seoul 08826, Korea (e-mail: jgkim2020@snu.ac.kr). Dohyun Jang is with the Department of Aerospace Engineering and the Automation and Systems Research Institute, Seoul National University, 1 Gwanak-ro, Gwanak-gu, Seoul 08826, Korea. (e-mail: dohyun@snu.ac.kr). H. Jin Kim is with the Department of Aerospace Engineering, Seoul National University, 1 Gwanak-ro, Gwanak-gu, Seoul 08826, Korea (e-mail: hjinkim@snu.ac.kr).

\textsuperscript{\P} These authors contributed equally to this work.

* Corresponding author.
}

\runningtitle{2022}{Jigang Kim, Dohyun Jang, and H. Jin Kim}{Manuscript Template for the International Journal of Control, Automation, and Systems: ICROS {\&} KIEE}{xxx}{xxxx}{x}

\section{Introduction}

\textcolor{red}{Multi-agent systems have garnered increasing attention recently within the robotics community. These systems have great potential to address practical applications in large-scale and unknown environments, including but not limited to search and rescue, reconnaissance, and surveillance \cite{alotaibi2019lsar,ma2016information,shi2020adaptive,wang2018master}. Specifically, we are interested in the challenge of multi-agent target search and tracking, which requires processing the combined signal intensity from multiple targets measured by agents at various locations and time intervals. In general, the signal intensity field is unknown beforehand in these situations \cite{lim2017bayesian}. Planning efficiently over the unknown signal field is not straightforward and leads to the exploration-exploitation dilemma \cite{srinivas2012information}. Reinforcement learning (RL) is a promising approach to intelligently deal with this dilemma by learning from past experiences in a data-driven manner \cite{sutton2018reinforcement}.}

However, there are multiple challenges to applying RL to a multi-agent system in unknown signal fields. While extending RL from single agent to multi-agent is straightforward in theory, it may be infeasible in practice. The observation-action space scales exponentially to the number of agents which hinders optimization (curse of dimensionality). Scalability issues aside, the number of agents or the connection between agents can vary in \textcolor{red}{multi-agent systems, which is difficult to accommodate with existing RL algorithms}. In addition, \textcolor{red}{partial-observability of the target search and tracking mission renders} it incompatible with RL, which assumes a fully-observable Markov decision process (MDP).

There exists a body of research on multi-agent reinforcement learning (MARL). These methods formulate the problem from a generalized Markov game perspective, such as decentralized partially-observable MDP (POMDP), and they assume a sparse interaction between agents to avoid the scalability issue\cite{yu2015multiagent,zhou2016multiagent}. This stems from the fact that each agent, in most cases, can rely only on its information and act independently of other agents unless coordination between agents is required. Thus, existing approaches mainly focus on partial-observability arising from agent-centric observation and sparse interactions rather than from the problem setting itself. However, this is not the case for distributed multi-agent target search and tracking, where partial-observability appears directly in the form of an unknown signal intensity environment.

In this paper, we propose to augment deep reinforcement learning (DRL) with distributed Gaussian process (GP) multi-agent consensus-based map building to tackle the partial-observability of the search and tracking mission. This hybrid approach transforms the original problem of POMDP into MDP by incorporating the belief of the environment map into the observation input of the RL agent. An added benefit of this approach is in combining the strengths of \textcolor{red}{GP and RL}, marrying the interpretability afforded by the GP with the \textcolor{red}{generalizability and transferability} of DRL algorithms. One of the downsides of RL is that it can be difficult to understand the factors that lead to the agents' actions, more so in a multi-agent setting. Thus, explicitly constructing a belief of the environment provides useful insight for diagnosing and refining the technique.

\subsection{Related works}
Various studies have tackled the problem of collecting environmental information using multiple robots equipped with sensors. In \cite{elwin2019distributed}, the finite element method (FEM) was used for persistent environmental monitoring with multiple robots. In \cite{yao2017gaussian}, multiple unmanned aerial vehicles (UAVs) generated a target probability distribution map using the Gaussian mixture model (GMM) for target search. A more common approach than these methods is to use GP. GP is a non-parametric regressor based on training data, which is advantageous for inferring spatial information using a kernel. In \cite{patrikar2020wind,binney2013optimizing,ma2016information}, the sensor data gathered from each robot was processed with GP, and multiple robots performed tasks such as wind map building and ocean monitoring. In \cite{luo2019distributed,shi2020adaptive}, environmental modeling was performed with a mixture of GP combining GMM and GP.

However, it is not straightforward how to extend conventional GP to distributed settings since GP requires that all data be collected and processed on a single device such as a server. Thus, decentralized techniques have been proposed to apply GP to distributed robot networks \cite{allamraju2017communication,pillonetto2018distributed,jang2020multi}. In \cite{pillonetto2018distributed}, a consensus algorithm was applied to a network of sensors to obtain an estimate of the entire map through local communications alone. In \cite{jang2020multi}, a multi-robot system performed environmental learning in a distributed network based on \textcolor{red}{the} distributed GP and consensus algorithm.

\textcolor{red}{Recent studies on multi-robot systems have integrated intelligent controllers with environmental learning algorithms to achieve higher autonomy.} One such approach is RL which lies at the intersection of approximate dynamic programming and stochastic optimal control. Recent advances in data-driven techniques such as deep learning have extended the reach of RL to various domains\cite{silver2017mastering,sun2018conversational,deng2016deep,choi2017inverse}. DRL has been successfully applied to complex skills and high-dimensional inputs both in simulation and physical systems\cite{gu2017deep,andrychowicz2020learning}. DRL may overcome the shortcomings of traditional methods in balancing the exploration-exploitation tradeoff of multi-agent target search and tracking mission by leveraging a diverse set of prior experiences to learn a generalized strategy that balances the tradeoff in an end-to-end manner.

Multi-agent multi-target capture has been considered in MARL domain\cite{omidshafiei2017deep}. However, such works mostly focus on the theoretical aspect and only consider grid-world examples which do not apply to robotic systems. Our work resembles previous work which applied MARL to UAVs\cite{qie2019joint}. It defined the multi-UAV target tracking problem and applied an existing MARL algorithm under the multi-particle simulation environment. Unlike prior work, we consider a more realistic setting where target information is initially unknown and must be gathered from exploration. To do so, we propose a GP-based cooperative map building and demonstrate our method not just in simulation but with multi-UAV experiments as well.

\subsection{Our contributions}
To achieve our objective, we focus on the following three main contributions:
\begin{itemize}
  \setlength\itemsep{-0.25em}
  \item[•] We build a belief for the target locations to circumvent the partial-observability arising from the unknown signal intensity field.
  \item[•] We decentralize the map building process with distributed GP, which improves the scalability of the system by alleviating the computational and network burden.
  \item[•] We modify an existing MARL algorithm to accommodate the belief map as an image input and facilitate policy transfer.
\end{itemize}
We perform various multi-agent target search and tracking simulations to evaluate the performance and transferability of our method and demonstrate on a swarm of micro-UAVs for experimental validation.

The outline of the paper is as follows. Section 2 briefly describes the problem statement and preliminaries. Section 3 details a method for cooperative map building and map-based MARL algorithm to tackle multi-agent target search and tracking mission. Simulations and multi-UAV experiments for the various scenarios are presented in Section 4. Section 5 concludes the paper.

\section{Problem statement and preliminaries}

\subsection{Problem statement}

We focus on the target search and tracking problem with $N$ multiple agents by considering a target domain as a 2-dimensional compact set $\mathbb{X}_c\subset\mathbb{R}^2$. There are $M$ targets deployed in the target domain, which locations are unknown in advance. Each agent has a sensor that measures signal power, which is radiated in all directions from each target. Since this signal power is inversely proportional to the travel distance, it can be a clue to estimating the target position. \textit{The main objective} is to locate all targets and position at least one agent for each target while avoiding agent-agent and agent-obstacle collision.

\subsection{Multi-agent systems}
We define the agent index set as $\mathcal{N}=\{1,2,\cdots,N\}$ for $N$ agents. The $i$-th agent's location at timestep $k$ is represented as $\mathbf{x}_{k}^{i}\in \mathbb{X}_c$ ($i\in \mathcal{N}$). Each agent can communicate with neighbor agents via peer-to-peer communication. We define a set of neighbors for robot $i$ \textcolor{red}{at time $k$} as $\mathcal{N}_{k}^{i}=\{j|\ {\lVert\mathbf{x}_{k}^{i}-\mathbf{x}_{k}^{j}\rVert}<d_{comm},j\in\mathcal{N}/\{i\}\}$, where $d_{comm}$ is the communication distance. $\mathcal{N}^{i+}$ is a shorthand for $\mathcal{N}^{i}\cup \{i\}$.

The target index set is represented as $\mathcal{M}=\{1,2,\cdots,M\}$ for $M$ multiple targets. Each target radiates a signal in all directions, and the signal intensity field $\varphi(\mathbf{x})$ at position $\mathbf{x}$ is defined by the sum of all signal intensities as follows:
\begin{equation}\label{eq:signal_intensities}
\renewcommand{\arraystretch}{1.2}
\setlength{\arraycolsep}{2pt}%
\begin{array}{ r>{{}}l @{\quad} l @{\quad} r>{{}}l @{\quad} l }
\varphi(\mathbf{x})=\sum\limits_{m\in \mathcal{M}}A^m h(\lVert \mathbf{p}^m - \mathbf{x} \rVert),\quad \mathbf{x}\in \mathbb{X}_c,\\
\end{array}
\end{equation} 
where $A^m\in \mathbb{R}_{\geq 0}$ and $\mathbf{p}^m\in \mathbb{X}_c$ are the maximum signal intensity and positions of $m$-th target, respectively. $h(\cdot)$ represents a monotonic-decreasing intensity function where $h(d)\rightarrow 0\ \textrm{as}\ d\rightarrow +\infty$.

Each agent has a signal power sensing unit that takes the measurement $y_k^i$ of $\varphi(\cdot)$ at its current position $\mathbf{x}_{k}^{i}$ ($i\in\mathcal{N}$), which has the following relationship:
\begin{ceqn}
\begin{equation}\label{eq:measurement_model}
\renewcommand{\arraystretch}{1.2}
\setlength{\arraycolsep}{2pt}%
\begin{array}{ r>{{}}l @{\quad} l @{\quad} r>{{}}l @{\quad} l }
y_{k}^{i}&=&\varphi(\mathbf{x}_{k}^{i})+n_{k}^{i},\\
\end{array}
\end{equation} 
\end{ceqn}
where the measurement of $\varphi(\mathbf{x}_{k}^{i})$ is corrupted by the additive white Gaussian noise $n_{k}^{i}\sim\mathcal{N}(0,\sigma_{n}^{2})$.

\subsection{Gaussian process for target search}\label{subsec:gaussian_process}

With the signal power measurement in \eqref{eq:measurement_model}, GP is utilized to estimate target locations. GP is a data-driven non-parametric regressor, which can provide probabilistic inference over the set $\mathbb{X}_c$, taking into account joint probability distribution between the sampled dataset \cite{rasmussen2003gaussian}. In \eqref{eq:measurement_model}, the unknown signal intensity field $\varphi(\cdot)$ is assumed to be represented as the following GP\textcolor{red}{:}

\begin{ceqn}
\begin{equation}\label{eq:process_model}
\renewcommand{\arraystretch}{1.2}
\setlength{\arraycolsep}{2pt}%
\begin{array}{ r>{{}}l @{\quad} l @{\quad} r>{{}}l @{\quad} l }
\varphi(\cdot)\sim \mathcal{G}\mathcal{P}(0,\kappa(\mathbf{x'},\mathbf{x''})),\\
\end{array}
\end{equation} 
\end{ceqn}
\textcolor{red}{where} $\kappa(\mathbf{x'},\mathbf{x''})$ is a \textit{kernel} for positions $\mathbf{x'},\mathbf{x''}\in\mathbb{X}_c$, and we use the original \textit{squared exponential} (SE) kernel which is defined as
\begin{ceqn}
\begin{align}\label{eq:kernel}
\kappa(\mathbf{x'},\mathbf{x''})=\sigma_{f}^2\exp(-\frac{1}{2}(\mathbf{x'}-\mathbf{x''})^\top\Sigma_{l}^{-1}(\mathbf{x'}-\mathbf{x''})),
\end{align}
\end{ceqn}
where $\sigma_{f}^2$ is the signal variance of $\varphi(\cdot)$, and $\Sigma_{l}$ is the length-scale matrix. The hyper parameters $\sigma_{f}^2$ and $\Sigma_{l}$ can be optimized by the maximum likelihood (ML) method \cite{rasmussen2003gaussian}.

When the agent $i$ samples the data $\mathbf{x}_k^i$ and $y_k^i$ at time $k$, these are stored sequentially in the GP input dataset $X_k^i=\{\mathbf{\bar{x}}_{1}^i,\cdots,\mathbf{\bar{x}}_{w_k^i}^i\}$ and the GP output dataset $\mathbf{y}_k^i=\{\bar{y}_{1}^i,\cdots,\bar{y}_{w_k^i}^i\}$, respectively. $w_k^i$ is the size of dataset at time $k$. For simplicity, we assume that $w_k^i$'s are identical for all agents, such that $w_k^i=w$ hereafter. For any test position $\mathbf{x}\in\mathbb{X}_c$, the posterior distribution over $\varphi(\mathbf{x})$ by agent $i$ is derived as follows:
\begin{ceqn}
\begin{equation}\label{eq:posterior_distribution}
p(\varphi(\mathbf{x})|X_k^i,\mathbf{y}_k^i,\mathbf{x}) \sim \mathcal{N}(\bar{\varphi}^i(\mathbf{x}),\Sigma^i(\mathbf{x})),\\
\end{equation} 
\end{ceqn}
where,
\begin{subequations}\label{eq:GP_mean_var}
\begin{align}
\bar{\varphi}^i(\mathbf{x})&=\mathbf{k}^\top(X_k^i,\mathbf{x})(\mathbf{K}(X_k^i,X_k^i)+\sigma_{n}^{2}I)^{-1}\mathbf{y}_k^i \label{eq:GP_mean_var(a)},\\
\begin{split}
    \Sigma^i(\mathbf{x})&= \kappa(\mathbf{x},\mathbf{x}) \label{eq:GP_mean_var(b)},\\
    &\ \ -\mathbf{k}^\top(X_k^i,\mathbf{x})(\mathbf{K}(X_k^i,X_k^i)+\sigma_{n}^{2}I)^{-1}\mathbf{k}(X_k^i,\mathbf{x})
  \end{split}
\end{align}
\end{subequations}
$\mathbf{K}(X_k^i,X_k^i)$ is $\mathbb{R}^{w\times w}$ matrix whose $(u,v)$-th element is $\kappa(\mathbf{\bar{x}}_u^i,\mathbf{\bar{x}}_v^i)$ for $\mathbf{\bar{x}}_u^i,\mathbf{\bar{x}}_v^i \in X_k^i$. $\mathbf{k}(X_k^i,\mathbf{x})$ is $\mathbb{R}^{w\times 1}$ column vector that is also obtained in the same way.

\subsection{Deep reinforcement learning}

RL is defined by a sequential decision making problem over MDP. MDP is a tuple of ($\mathcal{O}$, $\mathcal{A}$, $\mathcal{P}$, $r$), where $\mathcal{O}$ is the observation space, $\mathcal{A}$ is the action space, $\mathcal{P}$ is the transition probability $P(o_{t+1}|o_{t},a_{t})$, and $r(o_{t},a_{t})$ is the immediate reward function. The goal of a RL agent is to find the optimal policy $\pi^{*}(a|o)$ that maximizes the expected $\gamma$-discounted return $\mathbb{E}_{\pi}[\sum_{t=0}^{\infty} \gamma^{t}r(o_{t},a_{t})]$. The discount factor $\gamma \in [0,1]$ provides convergence guarantee for the Bellman operator and incentivizes the agent to take actions early.

With powerful function approximators, deep learning has pushed RL beyond the tabular setting. We opt for an off-policy, actor-critic variant of the DRL algorithm that can handle continuous observation and action spaces. Specifically, we opt for deep deterministic policy gradient (DDPG) algorithm\cite{lillicrap2016continuous} and its multi-agent variant MADDPG\cite{lowe2017multi}. Off-policy methods reuse past experience from the replay buffer $\mathcal{D}=\{(o_{t},a_{t},r_{t},o_{t+1})\}$ leading to better sample efficiency compared to their on-policy counterparts. Actor-critic methods consist of an actor which is a policy that maps observations to actions and a critic that evaluates the value of an observation-action pair under the given policy. The critic is updated via the temporal difference loss defined by the Bellman operator and the actor is updated by taking an ascending gradient on the critic.

\section{Method}

\begin{figure*}[ht]
\begin{center}
\includegraphics[trim = 0mm 0mm 0mm 0mm, clip, width=0.8\textwidth]{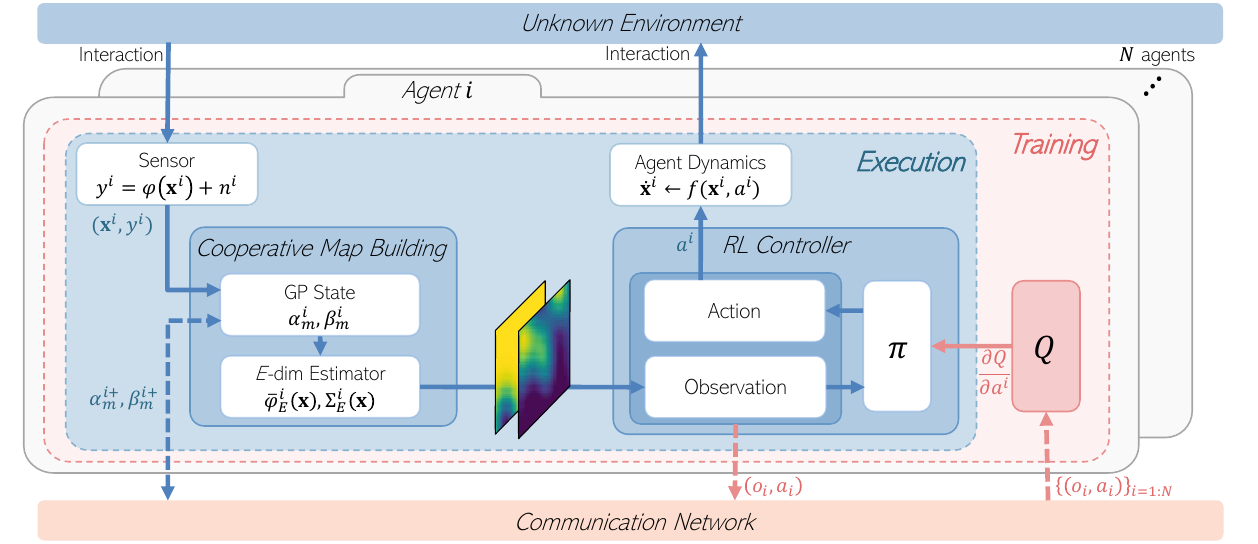} 
\caption{Schematics for target search and tracking with distributed GP and RL.} 
\label{fig:structure}
\end{center}
\vspace{-0.5cm}
\end{figure*}

The overview of multi-agent target search and tracking with distributed GP and MADDPG is shown in Fig. \ref{fig:structure}. $N$ agents, connected through a local communication network, operate while interacting with an environment with unknown targets. There are two main modules: \textit{cooperative map building} and \textit{RL controller}. Cooperative map building constructs a belief of the signal intensity map based on each agent's current position and sensor measurement. The actor network $\pi$ inputs the observation including the generated GP map and agent information to output the corresponding action. This action is then applied to the system and the observation is transitioned according to the agent dynamics and interaction with the unknown target environment.

RL controller is trained in a centralized manner and executed in a decentralized manner. During training, critic network $Q$ is evaluated with the observation-action pairs of all agents to update the actor network $\pi$. The reward function for the RL controller is a product of two sub-goal rewards, $r=r_{target} \times r_{collision}$. The sub-goal rewards are as follows:
\begin{subequations}
\begin{align}
    &r_{target} = \left[\prod_{m=1}^{M}0.1+0.9e^{-d^{m}/d_{char}} \right]^{\frac{1}{M}} \\
    &r_{collision} = \begin{cases}
    0, & \text{agent-agent/obstacle collision}.\\
    1, & \text{otherwise}.
  \end{cases}
\end{align}
\end{subequations}
Both sub-goal rewards are bounded to the range $[0,1]$. $r_{target}$ is additionally normalized by the number of targets. \textcolor{red}{$d^m$} is the distance between \textcolor{red}{$m$-th} target and its corresponding optimally assigned agent, derived from the linear assignment problem \cite{munkres1957algorithms}. $d_{char}$ is the environment-dependent characteristic distance that is set to some appropriate value. $r_{collision}$ is a binary reward that heavily penalizes any agent-agent or agent-obstacle collision\textcolor{red}{\sout{s}}.

\subsection{Multi-agent consensus-based map building}
Let the GP training dataset collected from all agents be $X_k=\cup_{i=1}^N X_k^i$ and $\mathbf{y}_k=\cup_{i=1}^N \mathbf{y}_k^i$. In the distributed multi-agent system, the conventional GP in Chapter \ref{subsec:gaussian_process} cannot incorporate $X_k$ and $\mathbf{y}_k$ due to the lack of a central server. For this reason, \textit{multi-agent consensus-based map building} is required.

The first step for consensus-based map building is to apply the $E$-dimensional approximation to the GP with Karhunen–Loève (KL) expansion \cite{levy2008karhunen}. This process is described in detail in \cite{jang2022fully}. According to $E$-dimensional approximation, the kernel \eqref{eq:kernel} can be represented in terms of eigenfunctions $\phi_e$ and corresponding eigenvalues $\lambda_e$ as $\kappa(\mathbf{x}',\mathbf{x}'')\approx \sum_{e=1}^{E}\lambda_{e}\phi_{e}(\mathbf{x}')\phi_{e}(\mathbf{x}'')$, and the $E$-dimensional estimator for \eqref{eq:GP_mean_var(a)} on the training dataset $X_k$ and $\mathbf{y}_k$ is expressed as follows \cite{pillonetto2018distributed}:
\begin{align}\label{eq:GP_E_dim_mean_estimator}
\bar{\varphi}_{E}(\mathbf{x}):=
  \Phi^\top(\mathbf{x})F_{E}\mathbf{y},
\end{align}
where,
\begin{subequations}\label{eq:GP_E_dim_mean_estimator_sub}
\begin{align}
\Phi(\mathbf{x})&:=
  \begin{bmatrix}
    \phi_{1}(\mathbf{x}),\cdots,\phi_{E}(\mathbf{x})
  \end{bmatrix}^\top,\\
F_{E}&:=
  \left(\dfrac{G^\top G}{wN}+\dfrac{\sigma_{n}^{2}}{wN}\Lambda_{E}^{-1}\right)^{-1}\dfrac{G^\top}{wN},\label{eq:GP_E_dim_mean_estimator_sub(b)}\\
G:=&
  \begin{bmatrix}
    \Phi(\mathbf{\bar{x}}_{1}^{1})\cdots\Phi(\mathbf{\bar{x}}_{w}^{1}),\cdots,\Phi(\mathbf{\bar{x}}_{1}^{N})\cdots\Phi(\mathbf{\bar{x}}_{w}^{N})
  \end{bmatrix}^\top.
\end{align}
\end{subequations}

The second step for consensus-based map building is to rewrite \eqref{eq:GP_E_dim_mean_estimator} into a distributed form, since constructing $G$ and $\mathbf{y}$ in \eqref{eq:GP_E_dim_mean_estimator} and \eqref{eq:GP_E_dim_mean_estimator_sub} requires centralized processing. The associated terms included in \eqref{eq:GP_E_dim_mean_estimator_sub(b)} can be decomposed as follows:
\begin{equation}\label{eq:GP_decomposition}
\dfrac{G^\top G}{Nw}=
\dfrac{1}{N}\sum\limits_{i=1}^{N}\alpha_{w}^{i},\quad
\dfrac{G^\top\mathbf{y}}{Nw}=
\dfrac{1}{N}\sum\limits_{i=1}^{N}\beta_{w}^{i},\\
\end{equation}
where $\alpha_{w}^{i}:=\sum\limits_{t=1}^{w}\Phi(\mathbf{\bar{x}}_{t}^{i})\Phi^\top(\mathbf{\bar{x}}_{t}^{i})/w$ and $\beta_{w}^{i}:=\sum\limits_{t=1}^{w}\Phi(\mathbf{x}_{t}^{i})\bar{y}_{t}^{i}/w$ are \textit{GP states} after the $w$-th sensor measurements.

\begin{figure*}[ht]
    \centering
    \includegraphics[width = 0.9\linewidth]{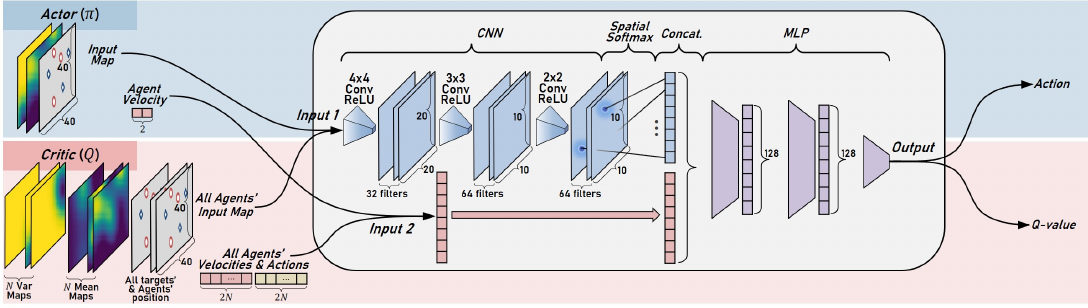}
    \caption{Network architecture for map-based MADDPG actor and critic.}
    \label{fig:network}
    \vspace{-0.5cm}
\end{figure*}

As a result, \eqref{eq:GP_E_dim_mean_estimator} is modified to the following distributed form:
\begin{ceqn}
\begin{equation}\label{eq:GP_E_dim_mean_estimator2}
\renewcommand{\arraystretch}{1.2}
\setlength{\arraycolsep}{2pt}%
\begin{array}{ r>{{}}l @{\quad} l @{\quad} r>{{}}l @{\quad} l }
\bar{\varphi}_{E}^{i}(\mathbf{x})&:=&
  \Phi^\top(\mathbf{x})\left(\alpha_{w}^{i}+\dfrac{\sigma_{n}^{2}}{wN}\Lambda_{E}^{-1}\right)^{-1}\beta_{w}^{i}.
\end{array}
\end{equation}
\end{ceqn}
Using the average consensus algorithm \cite{saber2003consensus} for the GP states, \eqref{eq:GP_E_dim_mean_estimator2} converges to \eqref{eq:GP_E_dim_mean_estimator} through repeated communication. Similarly, the distributed form of $\Sigma(\mathbf{x})$ in \eqref{eq:GP_mean_var(b)} is represented as follows:
\begin{ceqn}
\begin{equation}\label{eq:GP_E_dim_var_estimator2}
\renewcommand{\arraystretch}{1.2}
\setlength{\arraycolsep}{-1pt}%
\begin{array}{ r>{{}}l @{\ } l @{\ } r>{{}}l @{\ } l }
\Sigma_{E}^{i}(\mathbf{x}):&=&\kappa(\mathbf{x},\mathbf{x})-\Phi^\top(\mathbf{x})
\left(\alpha_{w}^{i}+\dfrac{\sigma_{n}^{2}}{wN}\Lambda_{E}^{-1}\right)^{-1}\\
&&\times \alpha_{w}^{i} \Lambda_{E}
\Phi(\mathbf{x}).
\end{array}
\end{equation} 
\end{ceqn}
When comparing distributed GPs with the conventional GPs in terms of \eqref{eq:GP_mean_var}, the computational complexity of the distributed algorithm is $O(E^{3})$, which is less than that of the centralized algorithm, $O((wN)^{3})$, since $E\ll wN$ holds in practice.

The final step in building the consensus map is to incorporate the newly acquired data into \eqref{eq:GP_E_dim_mean_estimator2} and \eqref{eq:GP_E_dim_var_estimator2} to maintain the continuity of GP estimate. Data is collected as each agent navigates the environment. Applying the online information fusion algorithm of \cite{jang2022fully}, the update rule of $\alpha_{w}^{i}$ and $\beta_{w}^{i}$ is defined as
\begin{ceqn}
\begin{equation}\label{eq:GP_alpha_beta}
\renewcommand{\arraystretch}{1.6}
\setlength{\arraycolsep}{2pt}%
\begin{array}{ r>{{}}l @{\ } l @{\ } r>{{}}l @{\ } l }
\alpha_{w+1}^{i}&=&\frac{w}{w+1}\alpha_{w}^{i}+\frac{1}{w+1}\Phi(\mathbf{\bar{x}}_{w+1}^{i})\Phi^\top(\mathbf{\bar{x}}_{w+1}^{i}),\\
 \beta_{w+1}^{i}&=&\frac{w}{w+1}\beta_{w}^{i}+\frac{1}{w+1}\Phi(\mathbf{\bar{x}}_{w+1}^{i})y_{w+1}^{i},
\end{array}
\end{equation} 
\end{ceqn}
where $\alpha_{0}^{i}=0$ and $\beta_{0}^{i}=0$. In the theorem 1 of \cite{jang2022fully}, it has been proven that new data are successively merged with the existing $\{\alpha_{w}^{i}\}_{i=1}^{N}$ and $\{\beta_{w}^{i}\}_{i=1}^{N}$, such that $\{\alpha_{w+1}^{i}\}_{i=1}^{N}$ and $\{\beta_{w+1}^{i}\}_{i=1}^{N}$ converge towards \eqref{eq:GP_decomposition} in a distributed manner.

\subsection{Map-based MADDPG}

We choose MADDPG as the base RL algorithm for the multi-agent target search and tracking mission. MADDPG overcomes the non-stationary dynamics arising from the partial-observability of multi-agent settings by adopting centralized critics $Q_{\psi_i}$ that inputs the combined set of observation and action spaces of all agents. MADDPG also proposes network ensembles for the decentralized actors $\pi_{\theta_i}$ to improve robustness and prevent the overfitting of policies with respect to other agents' actions which is especially problematic in competitive settings. To accommodate heterogeneous agents and individualized objectives, each agent \textcolor{red}{$i$} has a pair of centralized critic \textcolor{red}{$Q_{\psi_i}$} and decentralized actor \textcolor{red}{$\pi_{\theta_i}$}. This configuration also allows for centralized training and decentralized deployment. During training, the critic is updated by minimizing the temporal difference loss, and the actor is updated by gradient ascent on the critic. Temporal difference loss for the centralized critic of the $i$-th agent parameterized by network weights $\psi_{i}$ is as follows:
\begin{align}
    \hspace{-2mm}\mathcal{L}(\psi_i)=\mathbb{E}_\mathcal{D}[(Q_{\psi_i}^{\pi_{i}}(o^{1}_{t},\cdots,o^{N}_{t},a^{1}_{t},\cdots,a^{N}_{t}) - l^{i}_{t})^2],
\end{align}
where 
\begin{align*}
\hspace*{-3mm} l^{i}_{t}=r^{i}_{t}+\gamma Q_{\bar{\psi}_i}^{\pi_{i}}(o^{1}_{t+1},\cdots,o^{N}_{t+1},\pi_{\bar{\theta}_{1}}(o_{t+1}^{1}),\cdots,\pi_{\bar{\theta}_{N}}(o^{N}_{t+1})).
\end{align*}
The target q-value $l_{i}$ is evaluated with the target actor and target critic networks with delayed parameters $\bar{\theta}_{i}$ and $\bar{\psi}_{i}$.
Policy gradient for the $i$-th agent decentralized actor parameterized by network weights $\theta_{i}$ is as follows:
\begin{align}
    \hspace{-2mm}\nabla&_{\theta_{i}}\mathcal{J}(\pi_{\theta_{i}}) \\ &=\mathbb{E}_\mathcal{D}[\nabla_{a_{t}^{i}}Q_{\psi_i}^{\pi_{i}}(o_{t}^{1},\cdots,o_{t}^{N},a_{t}^{1},\cdots,a_{t}^{N})|_{a_{t}^{i}=\pi_{\theta_{i}}(o_{t}^{i})}\nabla_{\theta_{i}}\pi_{\theta_i}], \nonumber
\end{align}
where $\nabla_{a^{i}_{t}}$ denotes the action gradient of critic $Q_{\psi_i}$ with respect to the $i$-th agent's action. Note that only the $i$-th action is generated by the actor while the rest of the actions are sampled from the replay buffer $\mathcal{D}$.

We adapt the standard MADDPG implementation for the multi-agent target search and tracking task. Since our approach incorporates GP map as the input for the RL agent, we use convolutional neural nets (CNNs) to extract features from image inputs which are then fed into the multi-layer perceptron (MLP) to form the actor and critic networks (Fig. \ref{fig:network}). However, the combination of high-dimensional inputs with the multi-agent setting hinders RL. \textcolor{red}{Specifically, it hinders representation learning which captures and condenses relevant information from the high-dimensional inputs.} In practice, we make some modifications to aid representation learning. To reduce the number of network parameters, we adopt the spatial softmax module from \cite{levine2016end}. For typical CNNs, the \textcolor{red}{flatten layer} contains most of the network parameters. Spatial softmax replaces the \textcolor{red}{flatten layer} with a channel-wise soft-argmax operation that extracts 2D positions, applying a strong bottleneck without introducing additional parameters. In addition, it has been known to be beneficial to match the modality of various inputs \cite{chen2021cross}. For multi-agent target search and tracking mission, the agent has access to 1D agent-centric information such as relative position\textcolor{red}{\sout{s}} of other agents and obstacles as well as the GP map (Fig. \ref{fig:network}). Thus, we encode agent-centric information as an image and concatenate them alongside the GP mean and standard deviation channels forming a 3-channel image input for each agent.

The algorithm is further streamlined to improve computational efficiency. In the multi-agent target search and tracking mission \textcolor{red}{considered in this paper}, agents are homogeneous and share the same objective (reward function). Thus, we can share a single centralized critic \textcolor{red}{network} for all agents. We also share a single \textcolor{red}{actor network} for all agents. However, unlike sharing a critic which inputs the combined observations and actions of all agents, sharing an actor requires additional justification. Assuming homogeneous agents and shared objective, the optimal action for any given observation is invariant to the choice of actor: $a^{*}=\pi^{*}_{\theta_1}(o)=\cdots=\pi^{*}_{\theta_N}(o), \forall o \in O$. In other words, the actors of different agents will converge to the same optimal policy in theory ($\pi^{*}=\pi^{*}_{\theta_1}=\cdots=\pi^{*}_{\theta_N}$). Thus, we can share a single actor for all agents without the loss of optimality. A beneficial side-effect of this approach is the transferability to a varying number of agents during deployment. In standard MADDPG, there are multiple actors and it is not apparent which actor to exclude or duplicate when the number of agents decreases or increases during deployment. A single shared actor alleviates this ambiguity. We also do not use network ensembles for the shared actor. Network ensembles for actors are important in competitive settings but less so in cooperative settings\textcolor{red}{,} and policy robustness can still be maintained without ensembles since a single shared actor is trained with experience from all agents.

\section{Results}

\subsection{Hyperparameters}

\textcolor{red}{We conduct simulation and hardware experiments to evaluate the proposed method on multi-agent target search and tracking mission.} Simulation environment is based on the OpenAI multi-agent particle environment \cite{mordatch2018emergence} with \textcolor{red}{the} workspace of 2$\times$2. Radii of various entities such as the agent, target, obstacle were set to 0.05, 0.1, and 0.1. In addition, velocity and acceleration limit for the agents were set to 0.1 and 0.5, respectively. For signal intensities in \eqref{eq:signal_intensities}, $A^m=1$ was set for all $m\in\mathcal{M}$. The intensity function was set to $h(d)=\exp(-0.5d^2/0.06)$. \textcolor{red}{Likewise, the hardware experiment setup was configured to replicate the simulation environment but with a larger workspace of 4m$\times$4m.} 

The following hyperparameters were used throughout the experiments unless otherwise specified. For GP kernel \eqref{eq:kernel}, we set $\sigma_f^2=1$ and $\Sigma_l = \textrm{diag}([0.05, 0.05])$. The dimension $E$ of the approximated model was $40$. A communication range of 2 was used. The training environment included 3 agents, 3 targets, and 2 obstacles (A3T3O2) with randomized location\textcolor{red}{\sout{s}} of various entities at the start of every episode (domain randomization). \textcolor{red}{The} discount factor of 0.95 and \textcolor{red}{the} maximum episode length of 100 were used. Actor and critic networks in Fig. \ref{fig:network} were configured with three convolutional layers of (number of filters, kernel size, stride) tuple (32,4,2), (64,3,2), and (64,2,1) followed by two fully-connected layers of hidden layer dimension 128. The spatial soft-argmax module was applied to the last convolutional layer to extract a 128-dimensional feature vector which was then concatenated with agent velocity (decentralized actor) or velocities and actions of all agents (centralized critic) to be fed into the fully-connected network. A batch size of 1024 and a learning rate of 0.001 were used to update the networks every 100 steps.

\begin{figure}[t]
\includegraphics[width = 0.95\linewidth]{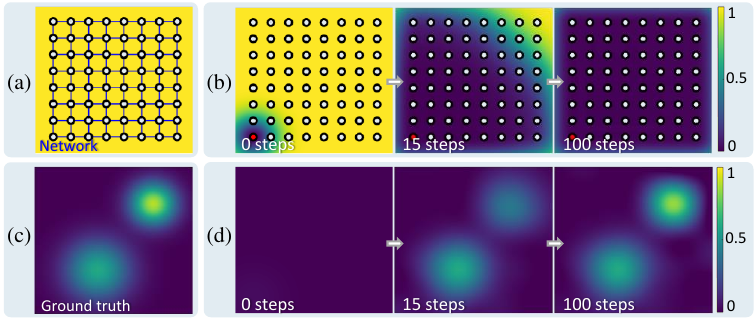}
\caption{An example of consensus-based distributed map building for multi-sensor array.}
\label{fig:gp_consensus_mean_var1}
\end{figure}

\subsection{Multi-sensor consensus-based map building}

In order to better visualize the inner workings of the consensus-based map building module under the distributed network environment, \textcolor{red}{it is applied to an example setting of a multi-sensor array.} In Fig. \ref{fig:gp_consensus_mean_var1}-(a), there are 64 uniformly distributed sensors throughout the map. They are connected with neighbors via local communication and each sensor measures only once at the beginning. Relying solely on local information exchange and \textcolor{red}{the} consensus algorithm with neighboring sensors, map information is shared with the entire network. Fig. \ref{fig:gp_consensus_mean_var1}-(b) shows the result of information propagation from the perspective of the agent in the lower left (red dot). Fig. \ref{fig:gp_consensus_mean_var1}-(c) is the ground truth data, and Fig. \ref{fig:gp_consensus_mean_var1}-(d) represents the process of building the map over time. As a result of the average consensus algorithm, all sensors' GP maps converge to the same global map.

\begin{figure}[b!]
    \centering
    \includegraphics[width = 0.83\linewidth]{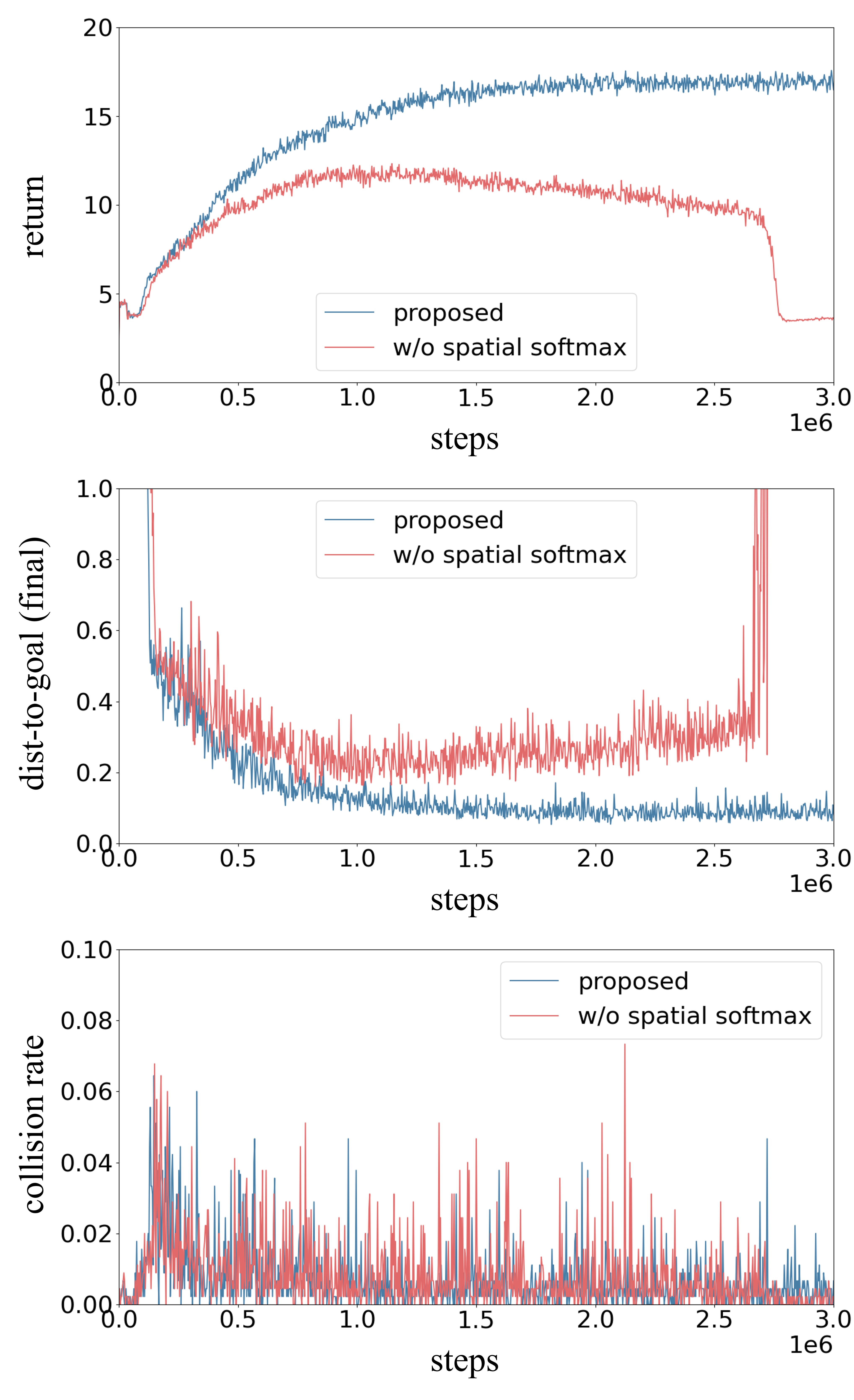}
    \caption{Network architecture ablation for map-based MADDPG.}
    \label{fig:cnn_ablation}
\end{figure}

\subsection{Map-based MADDPG}

To demonstrate the importance of representation learning in image-based MARL, we compare the learning curve of CNN with and without \textcolor{red}{the} spatial softmax module. \textcolor{red}{To eliminate the exploration burden on the MADDPG agent and test the effect of the spatial softmax module on representation learning in isolation, we assume that the cooperative map building module can recover and provide the MADDPG module with the ground-truth map.} Also, obstacles were removed from the training environment and the episode length was reduced for fast training and comparison. Fig. \ref{fig:cnn_ablation} compares the episodic return, distance-to-goal at final timestep, and collision rate of the proposed network and na\"ive convolutional neural net for 3M training steps.

In terms of the episodic return and distance-to-goal, the proposed network was up to 50\% more performant while the collision rate remained more or less identical for both cases. There was also a marked difference in learning stability as well, where the na\"ive CNN, in many training instances, has seen catastrophic failure occurs after convergence. This may be attributed to the overfitting of the actor due to a lack of network ensembles which is further compounded by \textcolor{red}{the} challenging representation learning under high-dimensional inputs.

\subsection{Simulation experiments}

We evaluate the performance of the proposed algorithm \textcolor{red}{(consensus-based map building + map-based MADDPG)} in simulation and check the transferability of the learned actor to a varying number of entities at test time. We consider the following three cases: varying number of agents and targets ($A=T=n$), varying number of targets ($T=n$), and varying number of obstacles ($O=n$). The first scenario assumes that the number of targets with unknown location\textcolor{red}{\sout{s}} is known and the matching number of agents can be deployed. In the second scenario, even the number of targets is unknown and a fixed number of available agents are deployed, most likely resulting in a mismatch between the number of agents and targets. The last scenario represents the varying threat level of the test environment. Various performance metrics for the three scenarios are shown in Table \ref{table:sim_results}. \textcolor{red}{$r_{avg}$ is the average episodic step reward, $d_{final}$ is the distance-to-target at the final timestep and $cr_{aa}$, $cr_{ao}$ is the agent-agent and agent-obstacle collision rate, respectively.} Note that the metrics for each configuration are averaged over 10 episodes with randomized entity locations.

\begin{table}[ht]
\caption{Performance metric for various configurations.}
\label{table:sim_results}
\begin{center}
\begin{tabularx}{1\linewidth}{c|*{4}{|X}}

\toprule

config. $^{*}$ & \centering $r_{avg}$ $^{\dagger}$ &\centering $d_{final}$ $^{**}$ &\centering ${cr}_{aa}$ $^{\dagger\dagger}$ &\centering ${cr}_{ao}$ $^{\dagger\dagger}$ \tabularnewline

\midrule

A3T3O2 & \centering 0.3648 &\centering 0.270 &\centering 0.025 &\centering 0.017 \tabularnewline

\midrule

\textcolor[RGB]{70,126,166}{A1T1}O2 & \centering 0.3781 &\centering 0.308 &\centering 0.000 &\centering 0.012 \tabularnewline
\textcolor[RGB]{70,126,166}{A2T2}O2 & \centering 0.3128 &\centering 0.324 &\centering 0.000 &\centering 0.006 \tabularnewline
\textcolor[RGB]{70,126,166}{A4T4}O2 & \centering 0.2875 &\centering 0.371 &\centering 0.075 &\centering 0.014 \tabularnewline
\textcolor[RGB]{70,126,166}{A5T5}O2 & \centering 0.2114 &\centering 0.334 &\centering 0.241 &\centering 0.016 \tabularnewline

\midrule

A3\textcolor[RGB]{70,126,166}{T1}O2 & \centering 0.3238 &\centering 0.257 &\centering 0.073 &\centering 0.029 \tabularnewline
A3\textcolor[RGB]{70,126,166}{T2}O2 & \centering 0.3414 &\centering 0.287 &\centering 0.043 &\centering 0.013 \tabularnewline
A3\textcolor[RGB]{70,126,166}{T4}O2 & \centering 0.3612 &\centering 0.267 &\centering 0.017 &\centering 0.006 \tabularnewline
A3\textcolor[RGB]{70,126,166}{T5}O2 & \centering 0.4110 &\centering 0.177 &\centering 0.003 &\centering 0.004 \tabularnewline

\midrule

A3T3\textcolor[RGB]{70,126,166}{O1} & \centering 0.3210 &\centering 0.341 &\centering 0.001 &\centering 0.004 \tabularnewline
A3T3\textcolor[RGB]{70,126,166}{O3} & \centering 0.3457 &\centering 0.291 &\centering 0.001 &\centering 0.023 \tabularnewline

\bottomrule

\end{tabularx}
\end{center}
\footnotesize{$^*$ \textcolor[RGB]{70,126,166}{blue font} denotes deviation from the training environment \\ $^\dagger$ average episodic step reward \\ $^{**}$ distance-to-target at the final timestep \\ $^{\dagger\dagger}$ agent-agent ('aa') or agent-obstacle ('ao') collision rate}
\end{table}

In the first scenario, $r_{avg}$ decreased as the number of agents and targets increased. Specifically, while performance was somewhat maintained for $d_{final}$, increasing rate of agent-agent collision negatively impacted $r_{avg}$. Since the map size has remained unchanged, increasing the number of agents increased the agent density, resulting in a more frequent collision between agents. For the special case of A1T1O2 with no agent-agent collision ($cr_{aa}=0$), $r_{avg}$ even outperformed the baseline. In the second scenario, the trends were reversed where the performance improved as the number of targets increased. Since the agent density remains a constant, $r_{avg}$ was directly tied with contention between agents. With fewer targets than agents, two or more agents fighting for the same target led to a higher chance of collision between agents. Varying the number of obstacles (scenario 3) did not meaningfully impact the result. There was a slight increase in $cr_{ao}$ as the number of obstacles increased but this did not have a meaningful impact on $r_{avg}$.

\begin{figure*}[b!]
    \centering
    \includegraphics[width = 0.99\linewidth]{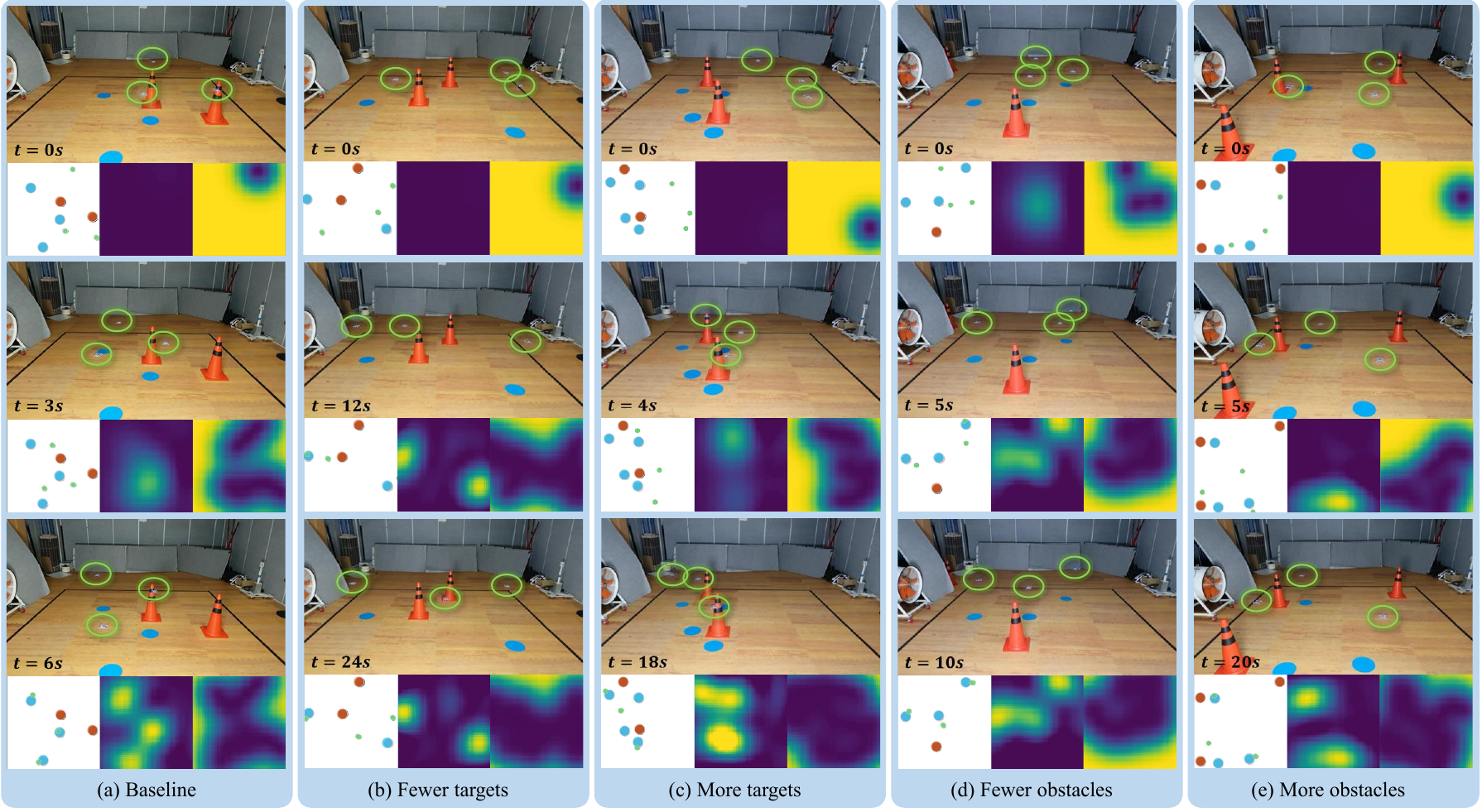}
    \caption{Multi-agent target search and tracking demonstration with micro-UAVs: each column represents a different \textcolor{red}{zero-shot} transfer setting with \textcolor{red}{a varying number of targets and obstacles. From top to bottom, snapshots of the episode are arranged in chronological order.} The three supplementary plots of each snapshot denote the top-down view, GP mean, and GP variance from left to right. \textcolor{red}{Note that zero-shot transfer is successful for minor($\pm1$) changes in the number of entities (targets, obstacles) and the UAVs were able to find and converge to unknown targets.}}
    \label{fig:exp1}
\end{figure*}

\subsection{Multi-UAV setup}

\begin{figure}[ht]
    \centering
    \includegraphics[width = 0.9\linewidth]{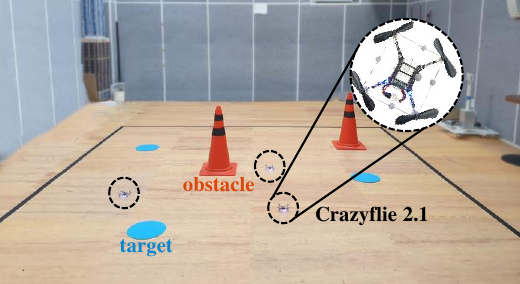}
    \caption{Multi-UAV experiment setup.}
    \label{fig:exp_setup}
\end{figure}

We demonstrate our algorithm with multi-UAV experiments. Fig. \ref{fig:exp_setup} represents the experiment setup for multi-agent target search and tracking mission. There are three targets (blue circle) and two obstacles (red cone) in a 4 m$\times$4 m workspace marked by black lines. Three Crazyflie 2.1 nano quadcopters measure the combined signal intensities of all targets. VICON motion capture system and \textcolor{red}{onboard} inertial navigation system\textcolor{red}{s} (INS\textcolor{red}{es}) were used for indoor localization. The entire system was implemented in Robot Operating System (ROS), and computations from the cooperative map building and RL controller modules were run on the laptop (Intel i7-7500U CPU) since Crazyflie 2.1 does not have an onboard microcomputer. The local communication links were implemented based on the distance between agents to mimic distributed control systems.

\subsection{Multi-UAV experiments}

Hardware experiments were performed on various transfer scenarios: one baseline scenario, two scenarios with varying numbers of targets, and two scenarios with varying numbers of obstacles. These scenarios only consider a subset of transfer scenarios presented in the simulation experiments due to time and physical constraints. Hardware demonstration was only performed once for each scenario.

Each snapshot in Fig. \ref{fig:exp1} displays \textcolor{red}{a captured footage and three supplementary plots (simplified top-down view, belief of the estimated target position denoted by the distributed GP mean and variance) at various times.} Indigo indicates low values and yellow indicates high values in GP plots. The result of the baseline experiment is shown in Fig. \ref{fig:exp1}-(a). For the GP variance plot, the uncertainty about the target gradually decreases as agents explore. In the case of GP mean plot, it is initialized to zero at the beginning of the episode and when an agent discovers a target, the estimated target location turns yellow. In summary, the performance was maintained for the baseline environment of A3T3O2 despite the domain gap between simulation and hardware. Considering that there are subtle differences in system dynamics between simulation and the real world, the performance degradation was minimal.

The hardware demonstration for reducing or increasing the number of targets is shown in Fig. \ref{fig:exp1}-(b),(c). Due to the domain gap, it took more time for the agent to converge to the target compared to the baseline scenario. When there were fewer targets than agents, two agents would converge to a single target or a surplus agent would keep searching for a non-existent third target. When there were more targets than agents, the agents would surround the targets or one agent would oscillate between two targets. In Fig. \ref{fig:exp1}-(d),(e), the hardware demonstrations for reducing or increasing the number of obstacles are shown. In the case of one obstacle, it is less difficult than the baseline and the agent would quickly converge to the target. In the case of three obstacles, it is more difficult than the baseline and took more time to find the target. \textcolor{red}{Overall, the trained agent could successfully zero-shot transfer to minor changes in the number of targets and obstacles.}

\section{Conclusion}

\textcolor{red}{This paper addressed the challenge of multi-agent target search and tracking under an unknown environment. The decentralized map building module was proposed to construct a belief map and locate unknown targets with a distributed network of agents. The network architecture for the MARL agent was devised to accommodate high-dimensional inputs to incorporate the belief map with RL and allow for a zero-shot transfer during deployment. The proposed method was validated with simulated and hardware experiments in various scenarios and was able to search and track multiple targets under an unknown environment whilst avoiding obstacles. We also evaluated the zero-shot transferability of the method to a varying number of environment entities during deployment and demonstrated that performance is maintained. These results indicate that the proposed algorithm is applicable to a broader class of autonomous multi-agent target search and tracking operations and can be adapted for practical applications.}

\section*{CONFLICT OF INTEREST}
The authors declare that there is no competing financial interest or personal relationship that could have appeared to influence the work reported in this paper.


\normalem

\bibliographystyle{ieeetr}
\bibliography{my_bib}

\biography{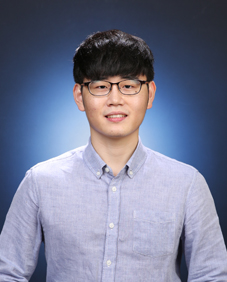}{Jigang Kim}{received his B.S. degree in Mechanical and Aerospace Engineering from Seoul National University in 2018. He is currently pursuing an integrated M.S./Ph.D. degree at the Department of Mechanical and Aerospace Engineering, Seoul National University. His research interests include robot learning, machine learning, and reinforcement learning.}

\biography{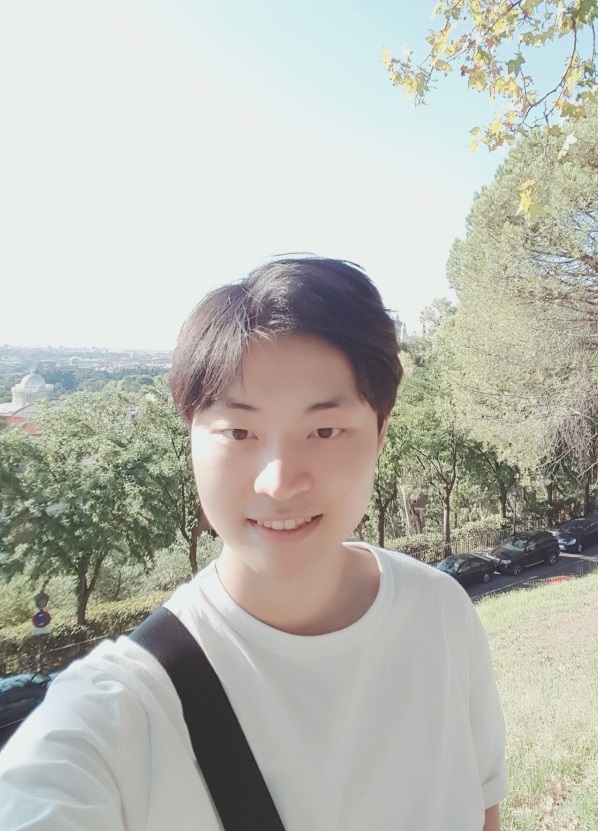}{Dohyun Jang}{received the B.S. degree in Electrical Engineering from Korea University in 2017, and the M.S. degree in Mechanical and Aerospace Engineering from Seoul National University, Seoul, in 2019. He is currently a Ph.D. Candidate in the School of Aerospace Engineering, SNU. His research interests include distributed systems, networked systems, machine learning, and robotics.}

\biography{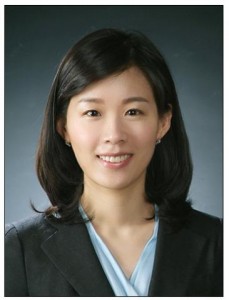}{H. Jin Kim}{received her B.S. degree from Korea Advanced Institute of Technology (KAIST) in 1995, and her M.S. and Ph.D. degrees in Mechanical Engineering from University of California, Berkeley (UC Berkeley), in 1999 and 2001, respectively. In September 2004 she joined the Department of Mechanical and Aerospace Engineering at Seoul National University, as an Assistant Professor where she is currently a Professor. Her research interests include autonomous robotics and robot vision.}


\clearafterbiography\relax

\end{document}